# The 2021 Urdu Fake News Detection Task using Supervised Machine Learning and Feature Combinations

Muhammad Humayoun

*Higher Colleges of Technology, Abu Dhabi, United Arab Emirates*

**Abstract**

This paper presents the system description submitted at the FIRE Shared Task: "The 2021 Fake News Detection in the Urdu Language". This challenge aims at automatically identifying Fake news written in Urdu. Our submitted results ranked fifth in the competition. However, after the result declaration of the competition, we managed to attain even better results than the submitted results. The best F1 Macro score achieved by one of our models is 0.6674, higher than the second-best score in the competition. The result is achieved on Support Vector Machines (polynomial kernel degree 1) with stopwords removed, lemmatization applied, and selecting the 20K best features out of 1.557 million features in total (which were produced by Word n-grams n=1,2,3,4 and Char n-grams n=2,3,4,5,6). The code is made available for reproducibility[1].

## 1. Introduction

As the world is getting more connected in the information age, fake news is also increasing. Spreading fake news is a proven tool in the propaganda warfare of the twenty-first century. Fake news can be spread to praise or defame an entity, person, group or society, based on geopolitical and religious motives. The methods and techniques for fake news detection are actively studied for major languages like English. Unfortunately, recourse poor languages are often neglected. In this context, Urdu fake news shared task[3] is an excellent step towards developing tools and techniques [1].

This paper presents the system description which was submitted at the competition. Our submitted results ranked fifth in the competition. Moreover, after the result declaration of the competition, we managed to attain even better results than the submitted results. The best F1 Macro score achieved by one of our models is 0.6674, which is higher than the second-best score in the competition. Some of the related research work outside of this competition describing the dataset construction and producing excellent results are reported in [2] [3].

Urdu is a widely spoken language in South Asia and worldwide due to the large South Asian diaspora [4]. Urdu has a modified Perso-Arabic alphabet, and it is written in cursive and context-sensitive Nastalique writing style. Urdu is unique because it takes its literary vocabulary from Persian and Arabic but informal vocabulary from the native languages of South Asia [5]. Some of the challenges that Urdu computing faces are: lack of capitalization, optional use of diacritic marks, and space not being a reliable word boundary marker [6] [7] [8]. In the absence of diacritics, context plays a vital role in guessing the pronunciation of a word. Urdu is a Subject-Object-Verb language having a free word order [9].

---

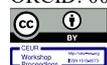

[3] https://www.urdufake2021.cicling.org/

## 2. Dataset Description

The dataset has 1,300 instances in the training set (including the test set within the training set). 750 instances are labeled as Real, and 550 instances are labeled as Fake. The test set has 300 instances (200 instances labeled as Real, 100 instances are labeled as Fake). The dataset is slightly imbalanced which could be ignored. A superficial analysis of the dataset reveals a very low number of non-standard script in the dataset. As expected in the news data, diacritic marks are absent. Data is generally clean. Proper segmentation of Urdu words remains an unresolved problem. However, tokenizing on spaces is the best strategy until proper word segmentation tools for Urdu are readily available.

Fake news detection is fundamentally a difficult problem. It is mainly because domain knowledge is needed to judge if news is fake or real. Anything that happens unexpectedly could be considered fake by those lacking sufficient domain knowledge. A recent example of this could be the fall of Afghanistan in the hands of the Taliban. It was such unexpected news that people felt the need to confirm it from more than one sources.

## 3. Preprocessing, Features and Classification Techniques

In supervised learning, the task of fake news detection can be modeled as a binary classification problem. A supervised learning algorithm known as a classifier is trained on a collection of training documents and their labels. Once training is completed, the classifier can take a document or text as an input and returns a label as an output. The framework we used consists of five steps: preprocessing (Section 3.1), feature extraction, and training classifiers (Section 3.2), producing labels, and their evaluation on reference labels (Section 4). For both tasks, train and test sets are given. The models are produced by training a classifier on a training set, and the label predictions are performed on a test set.

### 3.1 The Preprocessing

Preprocessing plays a key role in NLP. We apply the following preprocessing:
1. **Diacritic Removal**. Vowels are optionally used in Urdu. To ensure the consistency of data, removing all the vowels is a common practice.
2. **Text Normalization**. Persian and Arabic characters that visually look similar to their Urdu counterparts are sometimes used in writing, resulting in orthographic variations. We normalize all such variations to Normalization Form C [10].
3. **Stopword Removal**. The stopword list we used is provided by [7, 6], and it contains nearly 500 words.
4. **Lemmatization**. We used Urdu Morphological Analyzer [4] to convert all the surface forms of a word to its lemma or root. This tool covers approximately 5000 words, capable of handling 140,000 word forms.
5. **N-grams**. A list of tokens is produced by word and character n-grams (unigram, bigram, trigram, …).

### 3.2 Classification Techniques

Both classic supervised learning and neural network techniques have been extensively used in the literature for similar tasks [3] [11]. We have used the following two techniques:

#### 3.2.1 Support Vector Machines with K-Best features

We have used Support Vector Machines – SVM (Polynomial kernel degree 1) with K-Best features. One beneficial characteristic of SVM is the requirement of less memory to handle very large datasets. We have used this specific kernel and degree because of its better results in our initial experiments. A standard bag of words model is produced using a non-exhaustive list of features produced by character n-grams (n=2,3,4,5,6) and word n-grams (n=1,2,3,4). The value of features in a bag of word model is calculated using the TF-IDF weighting scheme. Since the number of features was huge, the K-best features were selected using *the SelectKBest* algorithm using Chi-squared statistic. Another reason to select K-best features was to keep a reasonable ratio between the number of features per instance and the total instances in the training set.

### 3.2.2 Convolution Neural Network

A Convolution Neural Network for sentence classification is reported in [11]. We used a simplified version of this model in which we have not used a pre-trained word Embedding. Pre-trained word Embeddings for Urdu such as this one [12] is available though we have not used it. It is mainly because, given the size of the task dataset, we were hoping to learn good embeddings from the dataset itself. We have used the following two variants:
1. The CNN model with four input channels. It was used for the reported results in the competition.
2. The CNN model with six input channels. During paper submission, we discovered comparable results with six input channels.

**The CNN Model:**

1. Each channel in the model is defined as:
   1.1. An input layer
   1.2. Embedding layer set to the size of the vocabulary and 100-dimensional real-valued vector.
   1.3. Convolutional layer of 1-dimension with 32 filters and a kernel size set to the number of words or characters to read at once (word or character n-grams where n=$k$ for channel$_k$ with k=1, 2, 3, 4, 5, 6. i.e. channel$_1$ used unigrams, channel$_2$ used bigrams, channel$_3$ used trigrams, and so on). Note that mixing of word n-grams and character n-grams are not possible.
   1.4. Max Pooling layer to combine the output from the convolutional layer.
   1.5. Flatten layer to reduce the three-dimensional output to two dimensional for concatenation.
2. The output from the six channels is concatenated into a single vector and processed by a Dense layer and an output layer. The model architecture with two channels for an example sentence is shown in Figure 1.

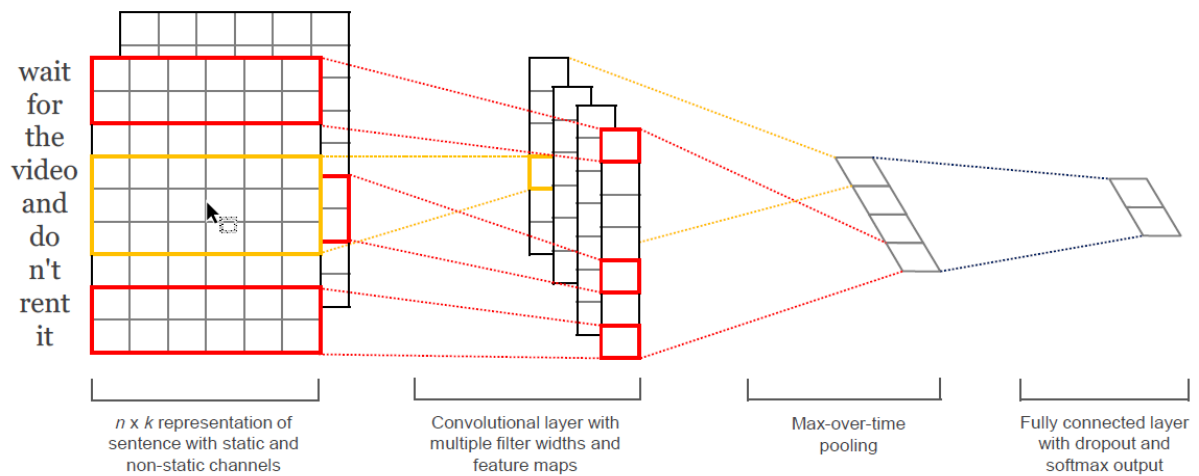

Figure 1: Model architecture with two channels for an example sentence taken from [11].

## 4. Experiments and Results:

For the task in hand, train and test sets are given. The models are produced by training a classifier on the training set, and the predictions are performed on the test set. The experiments are performed on a laptop with processor Intel Core i7 8th generation with 8 GB RAM.

### 4.1 Experiment 1

In this experiment, we produce a bag of word feature vector with a combination of word n-grams (n=1,2,3,4) and character n-grams (n=2, 3, 4, 5, 6). Note that stopwords are removed, and lemmatization is already applied in addition to the basic pre-processing settings mentioned in Section 3.1. The TFIDF weighting is applied to get the feature vector. The top results are shown in Table 1.

We learned that:
- The best score in Table 1, row 7 is better than the second best score in the competition.
- Excluding n=1 for char n-grams improve the results.
- The optimal number of features in K-Best is ~20K (see row 4 to 8).
- The best combination of features is: word n-grams n=1, 2, 3, 4 and char n-grams n=2, 3, 4, 5, 6.

**Table 1**
2021 Fake New Detection Task with **SVM Poly-1** and **K-best** Features. Stopwords removed and lemmatization applied. Best Score is **<u>bold faced and underlines</u>**, second best is <u>underlined</u>, third best is *italicized*.

| SN | K-Best | Fake Class | | | Real Class | | | F1 Macro | Accuracy |
|---|---|---|---|---|---|---|---|---|---|
| | | Prec | Recall | F1 Fake | Prec | Recall | F1 Real | | |
| | Word n-grams n=1, 2 and Char n-grams n=2, 3, 4, 5, 6 (total features: 1.177 million) | | | | | | | | |
| 1 | 20K | 0.5974 | 0.46 | 0.5198 | 0.7578 | 0.845 | 0.7991 | 0.6594 | 0.7167 |
| | Word n-grams n=1, 2, 3 and Char n-grams n=2, 3, 4, 5 (total features: 0.87 million) | | | | | | | | |
| 2 | 50K | 0.5542 | 0.46 | 0.5027 | 0.7512 | 0.815 | 0.7818 | 0.6423 | 0.6967 |
| 3 | 20K | 0.5647 | 0.48 | 0.5189 | 0.7581 | 0.815 | 0.7855 | 0.6522 | 0.7033 |
| | Word n-grams n=1, 2, 3, 4 and Char n-grams n=2, 3, 4, 5, 6 (total features: 1.557 million) | | | | | | | | |
| 4 | 70K | 0.5833 | 0.42 | 0.4884 | 0.7456 | 0.85 | 0.7944 | 0.6414 | 0.7067 |
| 5 | 50K | 0.625 | 0.4 | 0.4878 | 0.7458 | 0.88 | 0.8073 | 0.6476 | 0.72 |
| 6 | 25K | 0.5949 | 0.47 | 0.5251 | 0.7602 | 0.84 | 0.7981 | 0.6616 | 0.7167 |
| 7 | 20K | 0.6104 | 0.47 | 0.5311 | 0.7623 | 0.85 | 0.8038 | **<u>0.6674</u>** | 0.7233 |
| 8 | 10K | 0.6324 | 0.43 | 0.5119 | 0.7543 | 0.875 | 0.8102 | *0.661* | 0.7267 |
| | Word n-grams n=1, 2, 3, 4 and Char n-grams n=3, 4, 5, 6 (total features: 1.553 million) | | | | | | | | |
| 9 | 20K | 0.625 | 0.45 | 0.5233 | 0.7588 | 0.865 | 0.8084 | <u>0.6658</u> | 0.7267 |

### 4.2 Experiment 2

In this experiment, we performed a non-exhaustive list of the following constrained: (1) Number of channels to be 4, 5 and 6. Character level sequences and word level sequences (n-grams) through kernel size in the convolutional layer of each channel. On all of these experiments, stopwords were removed and lemmatization was applied in addition to the basic pre-processing settings mentioned in Section 3.1. Note that it is not possible to combine word n-grams and character n-grams in our implementation of CNN. It is mainly because we rely on the Keras tokenizer class which imposes the restriction of selecting if word sequences to be used or char sequences as a basic building block of the model. The results are shown in Table 2.

It can be seen that the results by CNN is inferior as compared to the results we achieved in Experiment 1. We think that it is mainly because of the mid-range size of the dataset. CNN models need massive amount of training instances in order to outperform traditional models. Such a dataset in our case is not available.

**Table 2**

2021 Fake New Detection Task with **Convolution Neural Network**. Stopwords removed and lemmatization applied. Best Score is **bold faced and underlines**, second best is underlined, third best is *italicized*. Results reported in the completion are shown in Row 1.

| SN | Fake Class | | | Real Class | | | F1 Macro | Accuracy |
|---|---|---|---|---|---|---|---|---|
|  | Prec | Recall | F1 Fake | Prec | Recall | F1 Real |  |  |
|  | **Character level CNN with 4 channels (with kernel sizes 1,2,3,4; one for each channel)** | | | | | | | |
| 1 | 0.48 | 0.49 | 0.49 | 0.74 | 0.74 | 0.74 | *0.611*[4] | 0.653 |
|  | **Word level CNN with 4 channels (with kernel sizes 1,2,3,4; one for each channel)** | | | | | | | |
| 2 | 0.49 | 0.58 | 0.53 | 0.77 | 0.7 | 0.73 | **0.629** | 0.656 |
|  | **Word level CNN with 6 channels (with kernel sizes 1,2,3,4,5,6; one for each channel)** | | | | | | | |
| 3 | 0.45 | 0.77 | 0.57 | 0.82 | 0.53 | 0.64 | 0.603 | 0.606 |
|  | **Character level CNN with 6 channels (with kernel sizes 1,2,3,4,5,6; one for each channel)** | | | | | | | |
| 4 | 0.47 | 0.66 | 0.55 | 0.79 | 0.64 | 0.70 | **0.627** | 0.643 |

## Conclusion

In this work, we have performed rigorous experimentation and achieved the second-best F1 Macro score of the competition. We demonstrated that traditional models with good feature engineering could produce good results for a mid-range dataset. In addition, the Neural Network-based methods such as CNN works reasonably well for the mid-sized dataset in hand. One way of improving the results by CNN might be the use of a pre-trained Urdu Embedding. However, such an investigation remains future work. Also, the recent transfer-learning techniques such as BERT fine-tuning can be investigated in future, though getting a large enough Urdu BERT model might be a challenge.

## References

[1]    M. Amjad, S. Butt, H. I. Amjad, A. Zhila, G. Sidorov and A. Gelbukh, "Overview of the shared task on fake news detection in Urdu at Fire 2021," in *In CEUR Workshop Proceedings*, 2021.

[2]    M. Amjad, N. Ashraf, A. Zhila, G. Sidorov, A. Zubiaga and A. Gelbukh, "Threatening Language Detecting and Threatening Target Identification in Urdu Tweets.," *IEEE Access,* vol. 9, pp. 128302-128313, 2021.

[3]    M. Amjad, G. Sidorov, A. Zhila, H. Gómez-Adorno, I. Voronkov and A. Gelbukh, ""Bend the truth": Benchmark dataset for fake news detection in Urdu language and its evaluation.," *Journal of Intelligent & Fuzzy Systems,* vol. 39, no. 2, pp. 2457-2469, 2020.

---

[4] These are the results reported in the competition.

[4]  M. Humayoun, H. Hammarstrom and A. Ranta, "Urdu morphology, orthography and lexicon extraction," in *In Ali Farghaly & Karine Megerdoomian (eds.), Proceedings of the 2nd Workshop on Computational Approaches to Arabic Script-based Languages. Pages 59–68, LSA 2007 Linguistic Institute, Stanford University, USA.*, California USA, 2007.

[5]  M. Humayoun, H. Hammarström and A. Ranta, "Implementing Urdu Grammar as Open Source Software," in *Conference on Language and Technology*, University of Peshawar, 2007.

[6]  M. Humayoun and H. Yu, "Analyzing Preprocessing Settings for Urdu Single-document Extractive Summarization," in *Proceedings of the Tenth International Conference on Language Resources and Evaluation ({LREC}'16)*, Portoroz, Slovenia, 2016.

[7]  M. Humayoun, R. M. A. Nawab, M. Uzair, S. Aslam and O. Farzand, "Urdu summary corpus," in *In Proceedings of the Tenth International Conference on Language Resources and Evaluation (LREC 2016) (pp. 796–800).*, Portoroz, Slovenia, 2016.

[8]  M. Humayoun and N. Akhtar, "CORPURES: Benchmark Corpus for Urdu Extractive Summaries and Experiments using Supervised Learning," *Intelligent Systems with Applications,* 2021.

[9]  M. Virk Shafqat, M. Humayoun and A. Ranta, "An open source Urdu resource grammar," in *Proceedings of the Eighth Workshop on Asian Language Resouces, 153-160*, 2010.

[10]  A. Gulzar, "Urdu Normalization Utility v1.0.," Technical Report, Center for Language Engineering, Al-kwarzimi Institute of Computer Science (KICS), University of Engineering, Lahore, Pakistan, 2007.

[11]  Y. Kim, "Convolutional Neural Networks for Sentence Classification," in *Empirical Methods in Natural Language Processing (EMNLP)*, Doha, Qatar, 2014.

[12]  S. Haider, "Urdu Word Embeddings," in *Proceedings of the Eleventh International Conference on Language Resources and Evaluation (LREC 2018)*, European Language Resources Association (ELRA), isbn:979-10-95546-00-9, Miyazaki, Japan, 2018.